%% file: main.tex
\documentclass[final]{cvpr}

\usepackage{times}
\usepackage{epsfig}
\usepackage{graphicx}
\usepackage{amsmath}
\usepackage{amssymb}

\usepackage{mathtools}
\usepackage{subfigure}

\usepackage{bbm}
\usepackage[ruled,vlined]{algorithm2e}

\def\@onedot{\ifx\@let@token.\else.\null\fi\xspace}
\def\eg{\emph{e.g}\onedot} 
\def\ie{\emph{i.e}\onedot}

\def\etal{\emph{et al}\onedot}

\newcommand{\Tref}[1]{Table~\ref{#1}}

\newcommand{\Fref}[1]{Fig.~\ref{#1}}

\newcommand{\Cref}[1]{Chap.~\ref{#1}}

\usepackage[pagebackref=true,breaklinks=true,colorlinks,bookmarks=false]{hyperref}

\def\cvprPaperID{2} 
\def\confYear{CVPRW MULA 2021}

\begin{document}

\title{Dealing with Missing Modalities in the Visual Question Answer-Difference Prediction Task through Knowledge Distillation}


\author{%
  Jae Won Cho\qquad%
  Dong-Jin Kim\qquad%
  Jinsoo Choi\qquad%
  Yunjae Jung\qquad%
  In So Kweon\\%
  KAIST, South Korea.\\ 
  \small{\texttt{chojw@kaist.ac.kr} \quad \texttt{djnjusa@kaist.ac.kr} \quad \texttt{jinsc37@kaist.ac.kr}}
  \\ 
  \small{\texttt{yun9298a@gmail.com} \quad \texttt{iskweon77@kaist.ac.kr} }
  }

\maketitle

\begin{abstract}

In this work, we address the issues of the missing modalities that have arisen from the Visual Question Answer-Difference prediction task and find a novel method to solve the task at hand. We address the missing modality--the ground truth answers--that are not present at test time and use a privileged knowledge distillation scheme to deal with the issue of the missing modality. In order to efficiently do so, we first introduce a model, the ``Big'' Teacher, that takes the image/question/answer triplet as its input and outperforms the baseline, then use a combination of models to distill knowledge to a target network (student) that only takes the image/question pair as its inputs. We experiment our models on the VizWiz and VQA-V2 Answer Difference datasets and show through extensive experimentation and ablation the performance of our method and a diverse possibility for future research.

\end{abstract}

\section{Introduction}
\input{1_sec_intro}

\label{sec.intro}


\section{Related Work}
\input{2_sec_related}


\section{Methodology}
\input{3_sec_method}


\section{Experiments}
\input{4_sec_experiment}

\hfill
\section{Conclusion}
\input{5_sec_conclusion}

{\small
\bibliographystyle{ieee_fullname}
\bibliography{egbib}
}

\end{document}

%% file: 1_sec_intro.tex

With the advancements of the Visual Question Answering (VQA)~\cite{antol2015vqa} task, where a model learns to generate a correct answer to a \emph{visual query} about a given image,
a new task called Visual Question Answer-Difference~\cite{bhattacharya2019does} (we call VQD for short) has been proposed where a model is required to take VQA one step further and try to understand why or how the answers of a VQA model may differ.

\begin{figure}[t]
\centering
    \resizebox{1\linewidth}{!}{%
    \includegraphics[width=\textwidth]{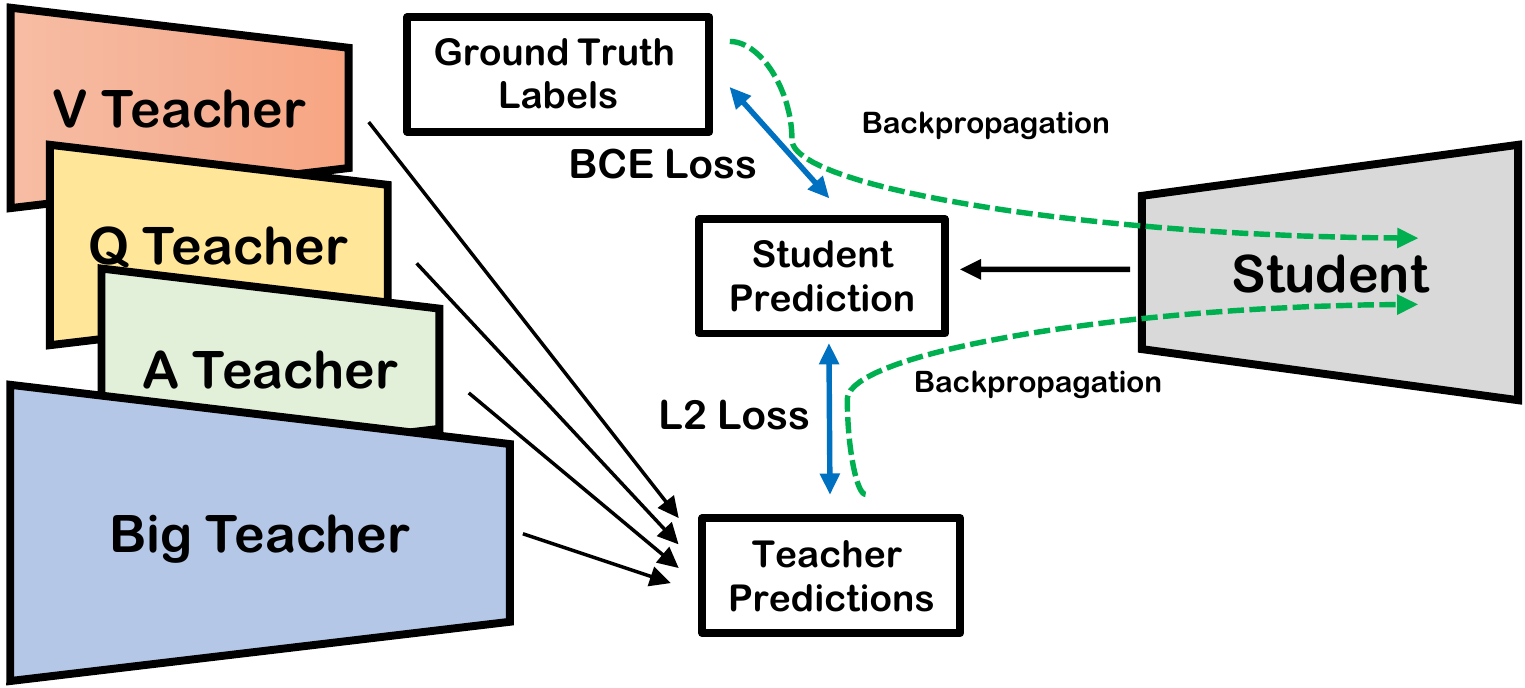}
    }
    \caption{Simplified illustration of Knowledge Distillation using Individual Modalities and a Combined Modality ``Big'' Teacher. The outputs of each teacher are leveraged and loss is backpropagated using L2 Loss into the Student Model along with the ground truth loss.}
\label{fig:teaser}
\end{figure}

The nature of the VQA task allows for an important real-world application of blind people using this framework to ask questions to an AI model to help them in their everyday lives~\cite{gurari2018vizwiz}.
However, with the limitations of current models, 
the designers of the VQD task~\cite{bhattacharya2019does} 
try to tackle a fundamental issue of the existing framework of models being trained to output the most probable answer without taking into consideration the subtle differences of answers or even the subjective nature of answers.

The VQD task is defined as follows: given a triplet of Image, Question, and the 10 possible Answers gathered through crowdsourcing, an AI model is required to answer out of 10 answer difference categories why the 10 possible answers might be different given the Image and Question.
Although this setting seems plausible, the creators of the task has not made the Answers of the Test set publicly available due to an ongoing challenge. This means that although we are free to use the Answers during training and validation, at test time, we are to use a different set of inputs.
To reiterate, in reality, the VQD task should be defined as follows: given a pair \emph{(not triplet)} of Image and Question, the model should be able to understand the intricacies of the Image and Question in order to output not only the correct answer but also the reasons as to why the answers could possibly be different. This makes the problem a much more challenging problem and also makes it much more real-world like.




Given this setting, we devise a method to tackle the challenge of the trying to understand the possibilities of the answers given only the Image and Question.
We first propose a network that uses all 3 modalities and outperforms previous networks. Then, we propose to use a knowledge distillation~\cite{hinton2015distilling} technique to distill knowledge about the missing modality to train another model that only has the Image and Question available to it. We show through our extensive experimentation and analysis the performance gains of this method on Vizwiz and VQA-V2 VQD dataset~\cite{bhattacharya2019does}. The main contributions of this work are summarized as follows:

\begin{enumerate}
    \item We revisit the Visual Question Difference (VQD) task~\cite{bhattacharya2019does} to show a novel method to deal with the realistic issue of the missing modality at test time.
    \item In order to effectively learn to predict VQD, we propose a new powerful network, ``Big'' Teacher, that attends to all three modalities and outperforms previous baselines. We believe the proposed model starts a new baseline for future research on VQD prediction task.
    \item We perform extensive analysis and experiments to show our novel method's intuitiveness and direction for possible future research in this task.
\end{enumerate}

%% file: 2_sec_related.tex
\label{ref:related}
\noindent\textbf{Visual Question Answering}
Recently, by virtue of the advancements in deep learning, there has been significant improvements in VQA models~\cite{lu2016hierarchical,schwartz2017high,yang2016stacked}, with recent VQA models mostly focusing on designing better visual/question bilinear fusion~\cite{ben2017mutan,fukui2016multimodal,gao2016compact,kim2016hadamard,yu2018beyond} or better attention mechanisms~\cite{anderson2018bottom,cadene2019murel,kim2018bilinear,lu2016hierarchical,yang2016stacked,yu2019mcan}.
However, the VQA task has shown several issues regarding its datasets such as
(1) Low dependency on visual cues~\cite{goyal2017making}, (2) unimodal bias~\cite{agrawal2018don}, (3) robustness~\cite{xu2018fooling}, and (4) answer differences~\cite{antol2015vqa}.
In this work, we specifically focus on the issue of answer difference in VQA datasets.

\noindent\textbf{Answer Difference in VQA Datasets}
As multiple annotators are involved in the labeling process for VQA datasets, visual questions often lead to different answers from different people~\cite{antol2015vqa,gurari2017crowdverge,malinowski2015ask}.
Even though we generally want to generate an answer that gives us the highest score for each visual question, a model trained on a datasets with an abundance of unique answers might generate ambiguous answers.
Previously, this problem has been addressed via a consensus based performance metric~\cite{antol2015vqa,malinowski2015ask}.
Recently,~\cite{bhattacharya2019does} introduced a dataset to explicitly learn why different answers occur.
The work by Bhattacharya~\etal extends prior works \cite{antol2015vqa,gurari2017crowdverge,malinowski2015ask} which has suggested reasons why answers can differ such as visual questions are difficult, subjective, ambiguous, or containing synonymous answers. 
In particular, they explicitly label each visual question indicating which among nine options are the reasons for the observed answer differences for two popular VQA datasets, VizWiz~\cite{gurari2018vizwiz} and VQA 2.0~\cite{goyal2017making}. The VizWiz VQD dataset, however, comes with an added challenge of the ground truth answers being missing at test time. We tackle this problem by using our novel training scheme with our new baseline.



\begin{figure*}[t]
\begin{center}
    \includegraphics[width=1\linewidth]{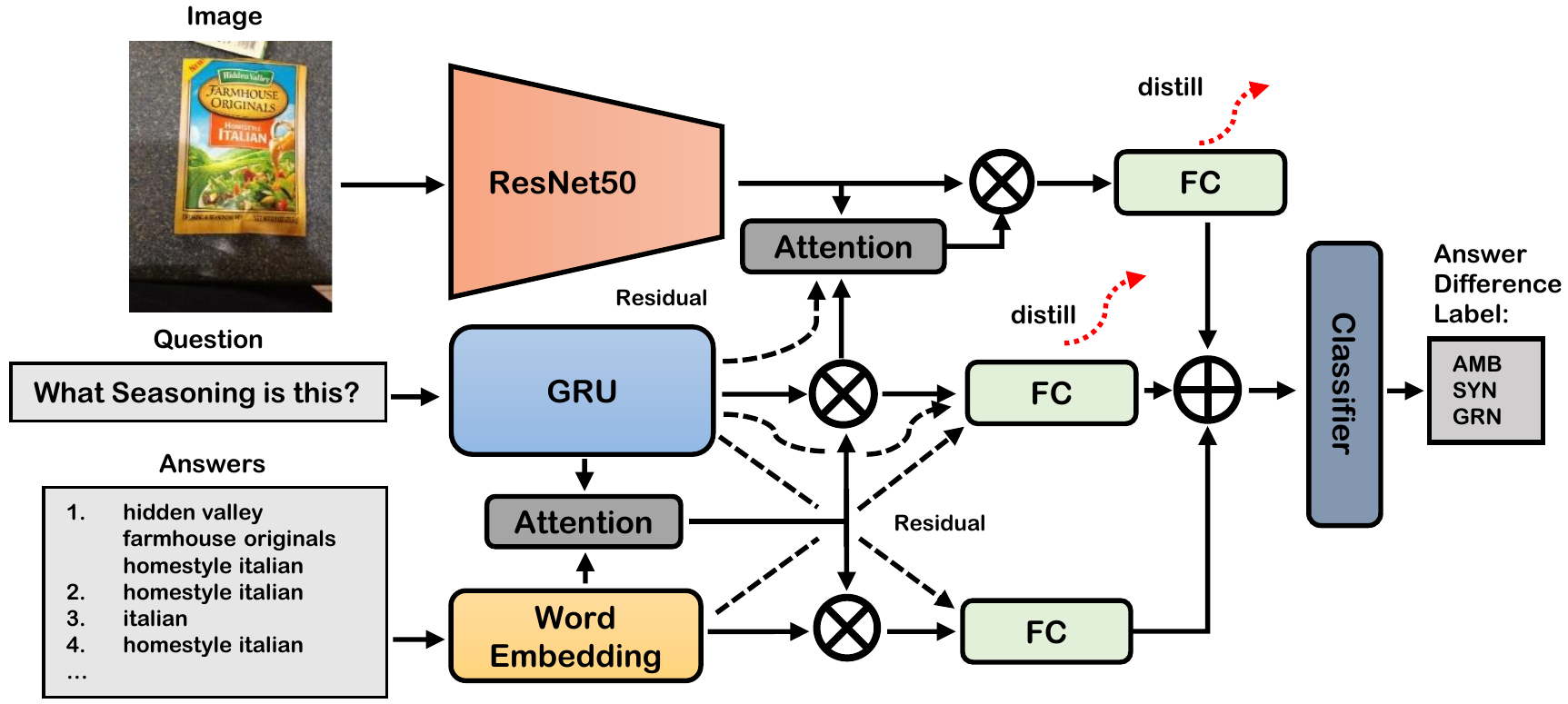}
\end{center}
    \caption{Architecture of our \emph{``Big'' Teacher} Model. We show with red dotted arrows the distilled features. Black dashed lines show residual connections.}
\label{fig:tripleatt}
\end{figure*}

\noindent\textbf{Generalized Knowledge Distillation}
Our proposed learning scheme is based on the concept of generalized distillation~\cite{lopez2015unifying} which combines the two popular ``machines-teaching-machines'' frameworks: knowledge distillation~\cite{hinton2015distilling} and privileged information~\cite{vapnik2009new}.
Vapnik and Vashist~\cite{vapnik2009new} first introduced a student-teacher analogy where they exploit a ``teacher'' model (model pre-trained with optical flow) that provides additional information about the training examples to a ``student'' model (model trained on RGB).
The teacher model is trained with additional information (\eg~depth map or optical flow) available only in the training phase and not at test time~\cite{vapnik2009new}, and they exploit the information of the teacher to better train the student. 
In our scenario, the ground truth answers are the privileged information available for training, along with visual/question, but only visual/question is available at test time.
On the other hand, Hinton~\etal~\cite{hinton2015distilling} introduced the concept of knowledge distillation.
The motivation stems from the desire to transfer knowledge learned from a teacher model (ensemble of models) to a student model (small model) by matching the outputs representation of the student to the outputs of the teacher in order to improve the performance of the student model.
In our case, we have several models that act as teacher models. More specifically, we have models for each individual modality (visual, question, and answer) and also a model that takes the visual/question/answer triplet as inputs while our target or student model only takes visual/question pair as inputs.
The generalized distillation approach is adapted by various applications that require cross-modality knowledge transfer including object detection~\cite{gupta2016cross,hoffman2016learning} and action recognition~\cite{crasto2019mars,garcia2019dmcl,garcia2018modality,kim2018disjoint,kim2020detecting,luo2018graph}.
To the best of our knowledge, we are the first to apply the generalized distillation scheme for VQD prediction task.

\begin{table}[ht]
\centering
\caption{The labels of reasons why answers differ defined by \cite{bhattacharya2019does}.}
\resizebox{.95\linewidth}{!}{%
\begin{tabular}{c || c}
    \hline
    \textbf{LQI} & Low Quality Image\\
    \textbf{IVE} & Insufficient Visual Evidence\\
    \textbf{INV} & Invalid Question\\
    \textbf{DFF} & Difficult Question\\
    \textbf{AMB} & Ambiguous Question\\
    \textbf{SBJ} & Subjective Question\\
    \textbf{SYN} & Synonymous Answers\\
    \textbf{GRN} & Granular (Answers present the same idea)\\
    \textbf{SPM} & Spam Answers\\
    \textbf{OTH} & Other\\
    \hline
\end{tabular}
}
\label{ansdifftable}
\end{table}

%% file: 3_sec_method.tex
In this section, we introduce the architecture of the teacher models utilized (including ``Big'' Teacher model) and the distillation framework we propose.

\subsection{Problem Definition}
The Visual Question Answering (VQA)~\cite{antol2015vqa} tasks is required to generate a correct answer $\hat{a}$ for a given visual question $(x,q)$.
The question $q$ is usually embedded into a vector with RNNs and the image $x$ is represented by fixed-size features from CNNs.
The traditional VQA model is trained with cross-entropy loss by comparing the model's predictions with ground truth answers $a$.



According to~\cite{bhattacharya2019does}, predicting visual question answer difference is designed as a multi-label classification problem.
Given an image and question pair $(x,q)$, the goal is to learn from the given binary ground truth vector $Y=\{y_i\}^N_{i=1}$ for each of the $N$ reasons ($N=10$ including ``Other'' class) why the answer might be different (the reasons are listed in \Tref{ansdifftable}).
For the ground truth labels, each class is labeled, either 1 (present) or 0 (not present), by five crowd workers.

While \cite{bhattacharya2019does} utilizes the image, question, and ground truth answers as input cues for their final model, as we are not provided with the ground truth answers for the testing phase, we can only rely on the image and question.
For this reason, our student model, which is our final model for answer difference predictions, only has image and question as inputs. Our final model utilizes ResNet50~\cite{he2016deep} for image features $x$, and a 300 dimensional pre-trained Glove word vector~\cite{pennington2014glove} followed by a single-layer GRU~\cite{chung2014empirical}.

Just like the given baseline for the model that makes use of the Question and Image (Q+I Model for short) as its input,  in \cite{bhattacharya2019does}, all the input cues (image representation $x$ and question representation $q$) are fed into our answer difference prediction model comprised of several fully connected layers and outputs the class logit $Z_p=\{\hat{z}_i\}^N_{i=1}$.
The final answer difference prediction $\hat{Y}=\{\hat{y}_i\}^N_{i=1}$ is computed via a Sigmoid non-linearity,
and the model is trained with the Binary Cross-Entropy (BCE) loss.


\begin{table*}[ht]
\centering
\caption{The notations of the distillation losses. We make use of 7 distillation losses in total by exploiting three teachers (Visual, Question, and ``Big'' Teacher) intermediary features and the final predictions.}
\resizebox{1\linewidth}{!}{%
\begin{tabular}{c|c|c|c|c|c|c}
    \hline
    $\mathcal{L}_{21}$&$\mathcal{L}_{22}$&$\mathcal{L}_{23}$&$\mathcal{L}_{24}$&$\mathcal{L}_{25}$&$\mathcal{L}_{26}$&$\mathcal{L}_{27}$\\\hline
    $\mathcal{L}_{L2}(Z^v_{vi},Z^s_{vi})$&$\mathcal{L}_{L2}(Z^q_{qi},Z^s_{qi})$&$\mathcal{L}_{L2}(Z^b_{vi},Z^s_{vi})$&$\mathcal{L}_{L2}(Z^b_{qi},Z^s_{qi})$&$\mathcal{L}_{L2}(Z^v_{p},Z^s_{p})$&$\mathcal{L}_{L2}(Z^q_{p},Z^s_{p})$&$\mathcal{L}_{L2}(Z^b_{p},Z^s_{p})$\\
    \hline
\end{tabular}
}
\label{table:loss}
\end{table*}

\subsection{Proposed Training Method}
As previously stated, for the VizWiz VQD dataset, the ground truth answers for the test split datasets are not available to the public. The task deals with two unknowns since both the ground truth answers, which are important inputs for the VQD task, and the answer difference labels are missing. This is due to both the VizWiz VQA task and VizWiz VQD task being separate challenges that are concurrently open on the Eval AI test server.

Because of this issue, we propose a new method so that we can use only the questions and the image to output the difference.
\cite{bhattacharya2019does,gurari2017crowdverge} claims that even without the answers, a model should be able to anticipate the different kinds of answers a question and image pair should have. 
In practice, however, this model does not perform as accurately as a model that has the ground truth answer as its input as this input contains significantly more information to the model. 
Intuitively, compared to a model that has all three modalities present, a model that has to rely on only the question and the image to guess why the answers \emph{could} be different would perform much worse as it is much more difficult.
In addition, since we do not have the implementation details of the Question, Image, and Answer (Q+I+A) Model given by~\cite{bhattacharya2019does}, we set our baseline model as the Q+I Model.

In light of this, we leverage a~\emph{teacher-student} framework with knowledge distillation~\cite{hinton2015distilling} to solve the current problem at hand. 
In this problem definition, we require several teacher models.
From Hinton \etal~\cite{hinton2015distilling} and intuitively, transferring more features and information generally leads to less overfitting and better performance compared to a model that lacks this extra information. 
Although the best performing models would have all modalities, with the current limitations of this task, we believe that using different combinations of teacher models is the best solution.

\subsection{Individual Modality Teacher Models}

As we propose that each modality gives important information to the student model when distilling knowledge based on what each modality learns, we describe in detail the teacher models.
The \emph{Visual Teacher} consists of a ResNet50~\cite{he2016deep} backbone to generate $Z^v_{vi}$, and we use $Z^v_{vi}$ through a simple Multi-Layer Perception (MLP) to generate class logit $Z^v_{p}$. The performance of this model is similar to that reported in \cite{bhattacharya2019does} as it follows the same architecture.
The \emph{Question Teacher} consists of a Glove~\cite{pennington2014glove} embedding and a GRU~\cite{chung2014empirical} backbone. Similar to the visual teacher, the embedded question $Z^q_{qi}$ is passed through a MLP to generate class logit $Z^q_{p}$. The performance of this model is also similar to that reported in ~\cite{bhattacharya2019does} for similar reasons.
We then propose a novel \emph{``Big'' Teacher}, a teacher model that outperforms the given baseline model with ground truth answer inputs and use this model to ultimately guide the student to make the correct predictions. 
This model uses all modalities and is the most complex model among all our models.

We also consider the \emph{Answer Teacher} that uses the diverse ground truth answers as input and tries to understand the reasons for this diversity. This is slightly different from the other teacher models in that it directly takes the answers and tries to guess why the answers would be different without any other cues. This is a naive approach and could be seen as counter-intuitive to use on its own, but we still test it to show its potential impact.
The official repository
uses a vocabulary of 6250 dimensions and this value is achieved only by taking answers that have appeared more than 5 times within the entire train/val/test dataset. The total number of unique answers that are given due to the issues of the open-ended crowd sourcing are given as 58,789~\cite{gurari2018vizwiz} and using all the unique answers might seem inefficient since there are too many values.
However, if the threshold is set at 5, too many unique answers are disregarded and the model may not be able to pick up on the minute differences that occur from either ambiguous or granular reasons. To this end, we recreate the answer vocabulary to include all available answers, from the train/val set, without the test set due to our lack of access, and set the vocabulary size to be 45,304. We also use this same vocabulary space for our \emph{``Big'' Teacher} model's answer input for the same reasons.

\begin{figure}[t]
\centering
    \resizebox{1\linewidth}{!}{%
    \includegraphics[width=1\linewidth]{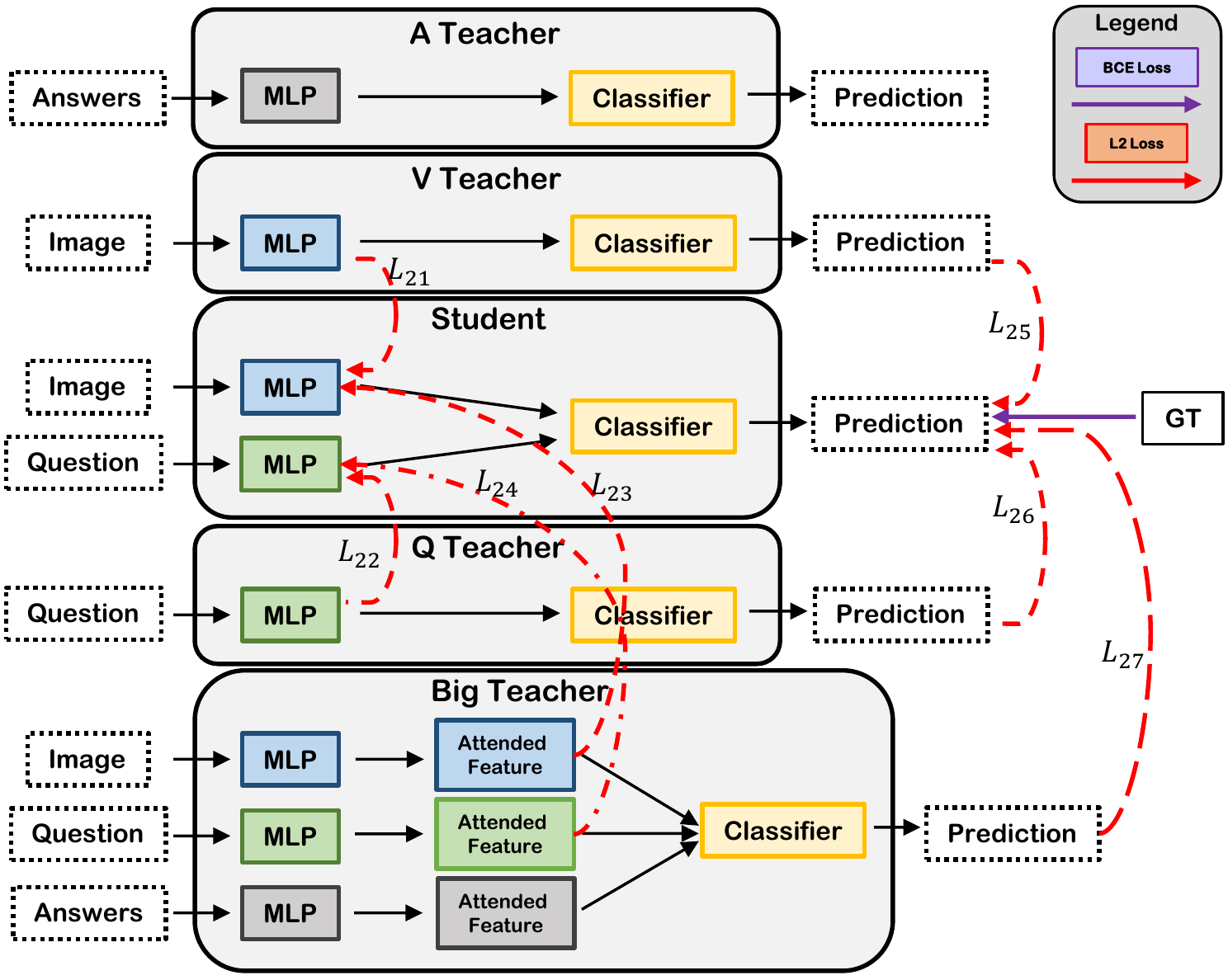}
    }
    \caption{Detailed model of our distillation method. The $L_{2i}$ listed correspond to the losses found in \Tref{table:loss}. Although A Teacher is shown in the figure, we do not use A Teacher in our final model, so we do not draw the distillation loss arrows}
\label{fig:intermeddistil}
\end{figure}

\subsection{``Big'' Teacher Model}
Our ``Big'' Teacher, as shown in \Fref{fig:tripleatt}, uses all modalities present, including the ground truth answers. 
The ``Big'' Teacher uses a ResNet50~\cite{he2016deep} backbone for image features and Glove ~\cite{pennington2014glove} embeddings with a GRU~\cite{chung2014empirical} for question features and uses the~\cite{anderson2017bottom}'s attention to attend to all three modalities.
Traditional attention simply attends the question to the image to see where the question needs to attend in the image. However, since we are dealing with three different modalities, we devise a different architecture suitable for attending three modalities.

First, we can attend the question to the answer, and thereby we can see whether or not the answer to the question is present.
At the same time, we can also attend the answer to the question to see if there is a part of the question that is important to the question. 
Attention goes both ways, so we attend it both ways as shown in \Fref{fig:tripleatt}.
For the question that has been attended by the answer, since it holds information about which part of the question is important to the answer, we attend this attended question to the image to see if there are parts of the image that correspond to this attended question. 
Additionally, in order to stop error propagation, we apply residual connections across the model.
We show that this model outperforms the given baseline significantly.
As a result, three feature vectors from three modalities are generated: $Z^b_{vi}$,$Z^b_{qi}$, and $Z^b_{ai}$ for visual, question, and answer features respectively.
The three feature vectors are summed up and pass through additional MLP to generate class logit $Z^b_{p}$ similar to multimodal fusion methods~\cite{kim2018bilinear,kim2019dense}.

\subsection{Student Model}


The \emph{Student Model} follows the same architecture of the Q+I model in~\cite{bhattacharya2019does} and consists of
a ResNet50 ~\cite{he2016deep} backbone with Glove~\cite{pennington2014glove} embeddings and a GRU~\cite{chung2014empirical} for its respective inputs. The features are passed through a FC layer to generate intermediary feature vectors ($Z^s_{vi}$ and $Z^s_{qi}$) then multiplies into a FC classifier for the output class logit $Z^s_{p}$. 
The final answer prediction from the student model $\hat{Y}^s$ is computed by feeding $Z^s_{p}$ into Sigmoid non-linearity.
A simplified \emph{Student Model} is shown in \Fref{fig:intermeddistil}.

\subsection{Teaching the Student}
Teacher Models need to be pre-trained in order to teach the student, so we train the teacher models with their respective inputs with BCE loss, 
then we use each of the Teacher Models as a guide in training the \emph{Student Model}. 
From each Teacher Model $t$ ($t\in\{v,q,b\}$), we make use of $L2$ loss to transfer the knowledge to the Student Model by matching the output:
\begin{equation}
    \mathcal{L}_{L2}(Z^t,Z^s) = ||Z^t-Z^s||^2_F. 
\label{L2loss}
\end{equation}

\begin{table*}[t]
\centering
\caption{The comparison between our ``Big'' Teacher Model (Ours) with the Baseline Models. Our ``Big'' Teacher Model shows the best performance among the baselines.}
\resizebox{.9\linewidth}{!}{%
\begin{tabular}{ l||c|cccccccccc }
    \hline
    &Overall&LQI&IVE&INV&DFF&AMB&SBJ&SYN&GRN&SPM&OTH\\
    \hline
    Q & 39.45 & 33.75 & 51.93 & 35.23 & \textbf{11.04} & 83.41 & 11.99 & 79.06 & 84.50 & \textbf{2.88} & 0.81 \\
    I & 40.50 & 56.69 & 50.44 & 26.98 & 7.58 & 83.18 & 9.44 & 80.27 & 86.40 & 2.12 & 1.91 \\
    A & 50.10 & 65.28 & 77.07 & 55.92 & 6.96 & 88.16 & 11.97 & 89.89 & 94.06 & 2.50 & 8.67 \\
    Q+I & 44.91 & 57.49 & 60.46 & 41.45 & 10.96 & 86.02 & 12.73 & 85.38 & 91.26 & 1.98 & 1.34 \\
    Q+I+A & 49.72 & \textbf{66.65} & 75.51 & 53.86 & 10.18 & 89.11 & 11.86 & 89.84 & 95.47 & 2.33 & 2.47 \\
    \textbf{Ours} & \textbf{51.60} & 66.63 & \textbf{77.98} & \textbf{57.74} & 9.7 & \textbf{89.14} & \textbf{12.89} & \textbf{90.63} & \textbf{95.79} & 2.06 & \textbf{13.46} \\
    \hline
\end{tabular}
}
\label{baselinetables}
\end{table*}

In addition to applying $L2$ loss on the predictions of the Teacher and Student Models, we also leverage the intermediary features of the Teacher Models, specifically from the \emph{Visual}, \emph{Question} and \emph{``Big'' Teacher}.
We use the intermediary visual and question features from the respective teachers and leverage $L2$ loss. These features are the features before the classifier and can be shown in \Fref{fig:intermeddistil}.

As single modality Teachers are trying to learn from their given modalities, 
each Teacher Model tries to leverage as much knowledge as it can from its given modality as there is no other modality to depend on. Due to this, if we force our Student Model to learn something from this feature representation, the Student Model can learn something that it normally would not due to the presence and dependence on another modality.

As for the \emph{``Big'' Teacher} Model, since the answer attends to the visual and question features, intermediary visual and question features hold information about the missing modality.
By distilling knowledge from these representation, the student model can learn to mimic the kind of representation it needs to follow.
By this logic, the Student Model learns to anticipate the kind of diverse answers that can exist from the question and image pair.

For our final model, we apply 7 $L2$ losses on the Student Model with one $BCE$ loss as shown below. 
We use the intermediary features and predictions of the \emph{Visual}, \emph{Question}, and \emph{``Big'' Teacher} to distill the loss to the \emph{Student} Model. The final loss equation is shown below:
\begin{equation}
    \mathcal{L}_{total} = \sum^{7}_{i=1}\lambda_i\mathcal{L}_{2i} + \lambda_0\mathcal{L}_{BCE}(Y,\hat{Y}^s),
\label{finalloss}
\end{equation}
where $\{\lambda_i\}$ and $\lambda_0$ are weights for losses with $L_2$ being $L2$ losses and $L_{BCE}$ being BCE loss.
The seven losses are a summation of individual losses and the detailed description of the seven distillation losses are shown in~\Tref{table:loss}.

In applying the loss in this form, the Student Model is able to understand diverse feature representations from each of the given modalities, allowing the Student to perform better that it could without this guide. A visual description is shown in \Fref{fig:intermeddistil} to aid in understanding the method at hand.

%% file: 4_sec_experiment.tex
In this section, we describe our experimental setups, experimental results, and implementation details.

\begin{table*}[t]
\centering
\caption{The performance of a Student Model with different combinations of Teacher Models with individual modalities, Visual (V), Question (Q), and Answer(A). Simply using single-modality Teacher Models is \emph{ineffective} in improving the Student Model. }
\resizebox{.9\linewidth}{!}{%
\begin{tabular}{ l||c|cccccccccc }
    \hline
    &Overall&LQI&IVE&INV&DFF&AMB&SBJ&SYN&GRN&SPM&OTH\\
    \hline
    Baseline & \textbf{44.91} & 57.49 & 60.46 & \textbf{41.45} & \textbf{10.96} & 86.02 & \textbf{12.73} & 85.38 & 91.26 & 1.98 & 1.34 \\
    A & 44.05 & 57.15 & \textbf{61.67} & 39.8 & 6.07 & 85.79 & 10.31 & 85.74 & 91.58 & 2.03 & 0.41 \\
    Q & 40.58 & \textcolor{red}{39.01} & 54.14 & 36.88 & 9.9 & 84.19 & 11.2 & 80.84 & 86.61 & 2.27 & 0.79 \\
    V & 40.96 & 55.68 & 52.13 & 28.89 & 7.76 & 83.77 & 9.32 & 81.72 & 87.15 & 2.02 & 1.15 \\
    V\&Q & 43.95 & 56.14 & 58.03 & 37.91 & 10.88 & 85.81 & 11.83 & 84.82 & 90.49 & 2.48 & 1.14 \\
    Q\&A & 43.99 & \textcolor{red}{53.12} & 59.65 & 41.1 & 8.31 & 86.47 & 11.85 & 85.66 & 90.88 & 2.32 & 0.59 \\
    V\&A & 44.45 & \textbf{58.92} & 60.85 & 38.84 & 7.81 & 85.99 & 10.55 & 85.45 & 90.82 & 2.29 & 0.99 \\
    V\&Q\&A & 44.68 & 56.83 & 60.06 & 39.27 & 10.02 & \textbf{86.92} & 12.02 & \textbf{86.14} & \textbf{94.41} & \textbf{2.46} & \textbf{1.69} \\
    \hline
\end{tabular}
}
\label{ablation}
\end{table*}

\subsection{Implementation Details}
As previously mentioned, we use a CNN backbone of ResNet 50~\cite{he2016deep} with a single layer GRU, and a modified answer space of 45,304.
We a word embedding to embed the answers into features instead of an RNN as the answers are not sentences. 
We set all hidden sizes to be 1024 throughout for all models present in our system including the attention layers.
Each MLP used in the models consists of 2 FC layers with a ReLU activation in between.
We use an Adam optimizer for all models with $\gamma = 0.1$ with a learning rate of $1e^{-2}$.
All teacher models are pretrained for 5 epochs while student models are trained for 20 epochs.

\begin{table*}[ht]
\centering
\caption{The performance of a Student Model with combinations of Teacher Models that includes our ``Big'' Teacher. (w/I) means the intermediary features. Note that combining all the Teacher Models (denoted as All) is not helpful. The Student Model trained using the ``Big'', Visual, and Question Teacher with all their intermediary features shows the best performance.}
\resizebox{1\linewidth}{!}{
\begin{tabular}{ l||c|cccccccccc }
    \hline
    &Overall&LQI&IVE&INV&DFF&AMB&SBJ&SYN&GRN&SPM&OTH\\
    \hline
    Baseline & 44.91 & 57.49 & 60.46 & 41.45 & 10.96 & 86.02 & 12.73 & 85.38 & 91.26 & 1.98 & 1.34 \\
    Big & 43.74 & 56.95 & 60.94 & 41.16 & 8.27 & 86.18 & 11.61 & 86.25 & 91.7 & 2.14 & 0.69 \\
    Big (w/I) & 44.61 & 57.51 & 50.88 & 40.2 & 9.97 & 86.4 & 11.34 & 85.71 & 91.14 & 2.38 & 0.55 \\
    Big\&Q & 44.23 & 51.56 & 60.61 & \textbf{41.74} & 10.65 & 85.78 & 12.15 & 85.8 & 91.03 & 2.02 & 0.9 \\
    Big\&Q (w/I) & 44.48 & 51.56 & 61.51 & 41.01 & 9.98 & 86.27 & 13.75 & 85.87 & 91.34 & \textbf{2.78} & 0.72 \\
    Big\&V & 45.36 & 58.01 & 61.09 & 39.12 & 9.71 & 86.26 & 12.2 & 85.54 & 91.1 & 2.06 & \textbf{8.53} \\
    Big\&V (w/I) & 44.67 & 58.27 & 60.42 & 38.02 & 10.5 & 86.67 & 10.37 & 85.38 & 90.98 & 2.32 & 1.74 \\
    Big\&A & 44.46 & 56.51 & 61.99 & 39.84 & 7.95 & 86.48 & 11.82 & 85.84 & 91.38 & 2.25 & 0.55 \\
    Big\&A (w/I) & 44.18 & 57.54 & 60.51 & 38.51 & 6.7 & 86.3 & 11.35 & 86.19 & 91.67 & 2.15 & 0.88 \\
    Big\&Q\&A & 44.25 & 54.9 & 61.98 & 40.77 & 6.81 & 86.63 & 11.39 & 85.91 & 91.11 & 2.45 & 0.53 \\
    Big\&Q\&A (w/I) & 44.05 & 55.07 & 60.12 & 39.6 & 7.49 & 86.23 & 11.0 & 86.08 & 91.63 & 2.3 & 0.97 \\
    Big\&V\&A & 44.23 & 57.98 & 61.19 & 38.06 & 7.85 & 86.6 & 11.02 & 85.6 & 91.17 & 2.07 & 0.75 \\
    Big\&V\&A (w/I) & 44.07 & 58.27 & 60.35 & 36.72 & 7.14 & 86.5 & 11.03 & 85.8 & 91.34 & 2.5 & 1.11 \\
    Big\&V\&Q & 45.41 & \textbf{59.09} & 61.49 & 39.57 & 11.91 & 86.47 & \textbf{14.08} & 86.23 & 91.66 & 2.08 & 1.5 \\
    Big\&V\&Q (w/I) & \textbf{45.75} & 58.46 & \textbf{62.39} & 39.87 & \textbf{12.71} & 86.52 & 11.52 & \textbf{86.31} & \textbf{91.85} & 2.32 & 5.52 \\
    All & 44.85 & 56.14 & 60.27 & 39.4 & 10.6 & \textbf{87.09} & 13.24 & 86.29 & 91.24 & 2.26 & 1.95 \\
    All (w/I) & 44.73 & 57.12 & 61.22 & 40.07 & 9.25 & 86.55 & 11.77 & 86.13 & 91.52 & 2.42 & 1.28 \\
    \hline
\end{tabular}
}
\label{ablationbig}
\end{table*}

\subsection{VizWiz VQD Dataset Setup}
We first test our method on the VizWiz VQD dataset~\cite{gurari2018vizwiz}, which consists of 19,176/3,063 (image, question, answers) triplets in the train/val splits respectively with additional 7,668 triplets for the test split. We only use the train set to train and validation set to test due to the test set not being publicly available
Each question is independently annotated with 10 answers and reasons for why answers differ, which has 10 classes with independent number of people annotating this separately. 
As mentioned above, we reconstruct a new vocabulary space for the answers from the train/val split.

\subsection{Answer Difference Baselines}
Since we use a different split from~\cite{bhattacharya2019does}, we set up our own baselines. 
\cite{bhattacharya2019does} does not give clear implementation details for the Question, Image, Answer (Q+I+A) Model, so we assume the Q+I+A\_GT~\footnote{Q+I+A\_GT is the model that takes the ground truth answers as input} Model as the Q+I+A Model from here on our for the purposes of this paper and show our comparisons in~\Tref{baselinetables}.

The Q, I, and A Models are models trained on individual modalities and are the models that we use as our teachers. 
The Q+I+A Model is the baseline model that \cite{bhattacharya2019does} proposes trained on our hyperparameters with the changed answer space, and the ``Big'' Teacher Model that we propose is denoted as \emph{\textbf{Ours}}.
As shown in \Tref{baselinetables}, Q Model shows high performance in question related classes (\ie INV, DFF, SBJ), and I Model shows high performance in image related class (\ie LQI).
As answers hold essential information to reason the answer difference, the A Model shows higher scores compared to Q Model and I Model in almost all classes and especially so in the Other category.
Combining question and image information leads to performance improvements (Q+I Model), and adding answer information further improves the performance.
Finally, we show from our ``Big'' Teacher Model that attending all three modalities with our method further boosts the performance. 
This more powerful model is shown to be crucial in distilling knowledge to the student model.

\begin{table*}[ht]
\centering
\caption{The Baseline Models for the VQA 2.0 Answer Difference Dataset.}
\resizebox{.9\linewidth}{!}{%
\begin{tabular}{ l||c|cccccccccc }
    \hline
    &Overall&LQI&IVE&INV&DFF&AMB&SBJ&SYN&GRN&SPM&OTH\\
    \hline
    Q & 43.61 & 8.04 & 58.62 & 44.25 & 28.67 & 96.74 & 23.99 & 89.1 & 85.02 & 1.33 & 0.36 \\
    I & 31.52 & 3.93 & 29.24 & 8.55 & 16.96 & 92.52 & 17.29 & 73.81 & 72.12 & 0.55 & 0.2 \\
    A & 42.57 & 9.62 & \textbf{59.76} & \textbf{46.86} & 18.94 & 96.49 & 17.3 & 89.63 & \textbf{85.81} & 0.91 & \textbf{0.42} \\
    Q+I & 43.19 & 8.32 & 57.27 & 41.32 & 28.98 & 96.35 & 24.87 & 88.31 & 83.49 & \textbf{2.72} & 0.34 \\
    Q+I+A & 43.2 & 8.57 & 57.52 & 40.33 & 27.88 & 96.73 & \textbf{26.12} & 89 & 84.67 & 0.87 & 0.33\\
    \textbf{Ours} & \textbf{44.32} & \textbf{10.39} & 59.73 & 43.22 & \textbf{31.5} & \textbf{96.89} & 24.19 & \textbf{89.73} & 85.42 & 1.75 & 0.39 \\
    \hline
\end{tabular}
}
\label{baselinetables_vqa2}
\end{table*}


\begin{figure*}[ht]
\centering
    \includegraphics[width=1\linewidth]{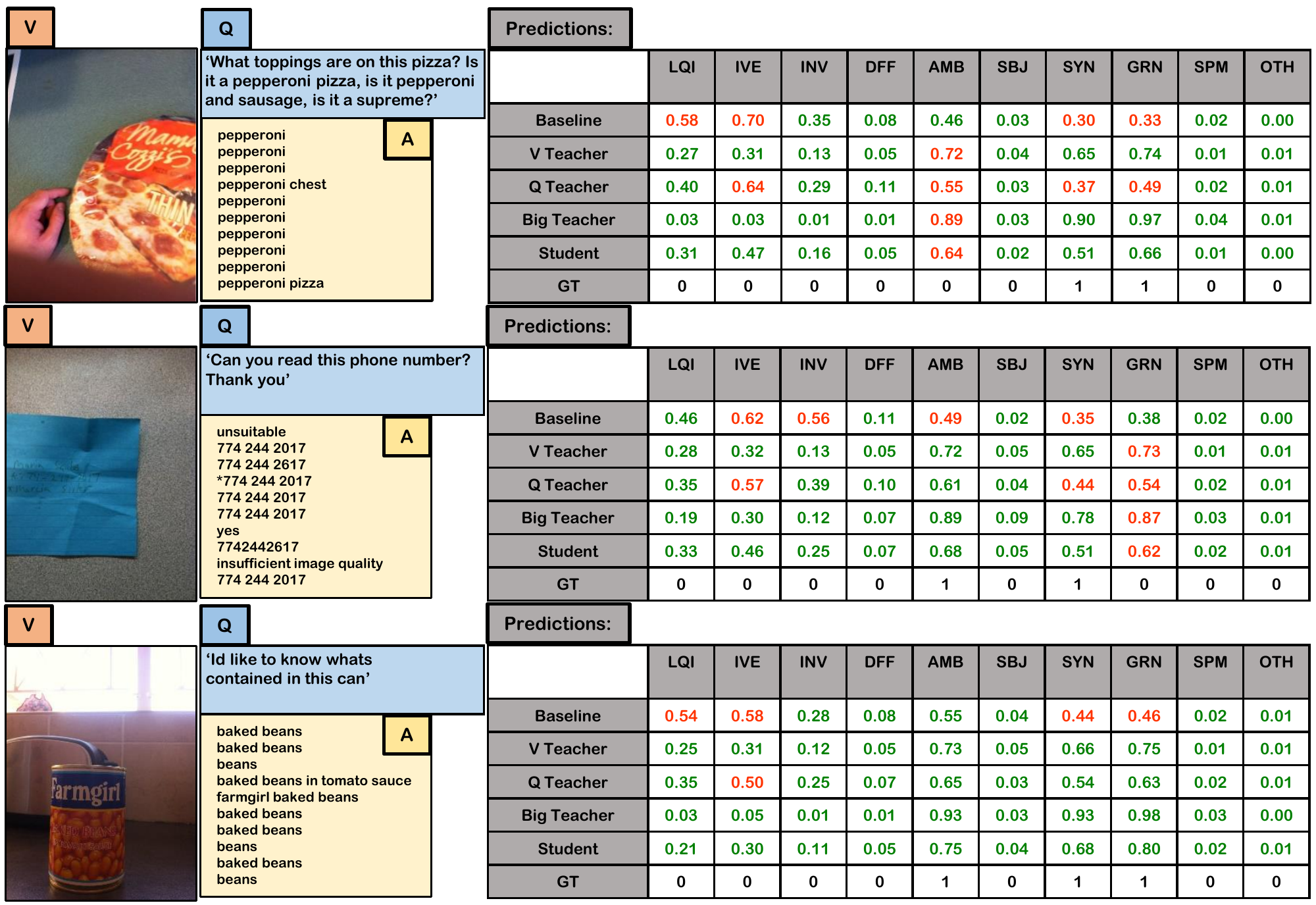}

    \caption{Qualitative examples showing how the Teacher Models affect the Baseline Model, and how it changes its answers after distillation. The output probability predictions are computed via Sigmoid function, and we hold the probability threshold of 0.5 to be a positive prediction. The numbers in red show wrong answers and green show correct answers. While the Baseline Model has wrong predictions, by virtue of the Teacher Models, our Student Model is able to generate correct predictions.}
\label{fig:qualitative}
\end{figure*}

\subsection{Results of Knowledge Distillation}
In light of the baseline models and the performance of our ``Big'' Teacher Model, we conduct extensive experimentation to see how each distillation method would affect the performance of the student model.
``Baseline'' is the Q+I Model without any knowledge distillation loss.
We evaluate all possible teacher combinations of A Model, Q Model, and V Model including intermediate feature losses denoted as ``w/I''.
We list our findings in \Tref{ablation}.

\Tref{ablation} show that distilling individual modalities into the baseline does not necessarily
aid in the student model, but shows that it can be counter productive to the model. 
For example, in \Tref{ablation}, we can see that not having any visual ques from the teachers would actually harm the model in understanding of the image such as LQI (Low Quality Image) as shown in red. 
This clearly shows the degradation that can occur from the individual modality teachers.
Generally for the criterions that are easy regardless, \emph{AMB, SYN, GRN}, it shows that the difference is minuscule.
Although the Answer Teacher performs significantly better than the Question and Visual Teachers, the Student Model does not seem to gleam too much knowledge from the Answer Teacher. This could be due to the fact that the Answer Teacher's predictions are based on a different modality that is not present in the Student Model.
Here, we show that the individual modality teachers are not helpful for the Student Model.

To combat this, we further experiment with distilling knowledge from the ``Big'' Teacher Model and show our ablations in \Tref{ablationbig}.
Through extensive ablation as shown in \Tref{ablationbig}
we find the most powerful combination is to ``Big'', Visual, and Question Teacher with all their intermediary features.
The intermediary features from the ``Big'' Teacher gives a guide to the Student Model's features as it has rich representations embedded in it from attention.
In addition, although the ``Big'' Teacher's visual and question features are rich, the ``Big'' Teacher is prone to being overly dependent on the answers, and we can see this through the \emph{Big} metric that it still does not outperform the best model. 
To combat the ``Big'' Teacher's reliance on the ground truth answers, we apply the losses of the Visual and Question Teacher and show that 
the Visual and Question Teachers' intermediary features
aids in boosting performance significantly. 



\subsection{Testing on the VQA 2.0 Dataset}

The VQA 2.0 Answer Difference Dataset has a relatively smaller dataset size of 15034, (train/val/test: 9735/1511/3788) compared to the 29,907 in the VizWiz Answer Difference Dataset.
The VQA 2.0 Dataset has the ground truth answers available for the test split, thus we use the trainval split to train the models and the test split for evaluation.

Generally, the trend of the classes that are easily wrong are different from that of the VizWiz Dataset. \Tref{baselinetables_vqa2} shows that the VQA dataset shows to have a much lower LQI (Low Quality Image) score as there are few number of such images within the dataset unlike the VizWiz Dataset. For the answers that are easy on the VizWiz set, \ie~AMB, SBJ, and GRN (Ambiguous, Synonym, and Granular), we see that the model's performance is not hugely affected. On the other hand, it seems that the classes such as DFF and SBJ (Difficult and Subjective) show a relatively higher score.

By looking at the Visual Teacher, or the standalone Visual Model, on the VQA dataset, it is very difficult to find any answer differences just by looking at the image.
This idea is transferred on to the classifications of the classes as we see the clear performance drop in the class LQI.

Surprisingly, in the ablation study on VQA 2.0 dataset as shown in \Tref{ablation}, the individual modalities noticeably improve the model's performance.
However, the best performing model still requires the ``Big'' Teacher Model as shown in \Tref{ablationbig}.

\Fref{fig:qualitative} shows qualitative results of how the Teacher Models affect the Baseline (Q+I) Model's performance. 
The figure shows the changes in predictions that the Student Model makes after having knowledge distilled from the Teacher Models. 
The model named ``Baseline'' is the student model without knowledge distillation, and the model named ``Student'' is our proposed model which learns through knowledge distillation.
In most of the cases, although the Baseline generates wrong predictions, by learning from teachers that can generate correct outputs, the student model is also able to generate correct predictions.
One drawback is that if the Teachers are all wrong, the Student can also be affected by this and change from a correct answer to a wrong answer (\emph{e.g.} AMB class in the first example and the GRN class in the second example). 
This shows the importance of a plausible Teacher Model when distilling the knowledge.

%% file: 5_sec_conclusion.tex
In this paper, we revisit the task of Answer-Difference Prediction with Visual Question Answering.
We study 
how to deal with the realistic problem of ground truth answers not being available at test time.
First, we devise a method to improve the performance of a model with the given modalities.
Through the use of all available modalities, we train individual teacher models and a plausible ``Big'' Teacher model. 
Our ``Big'' Teacher model shows favorable performance against the current baselines and sets a new bar for future works to come. 
Ultimately, we use the given teacher models to distill the information into a Student model with the privileged information they are given as our solution for our given problem. 
We believe this work can be the new first step towards solving VQD task and can inspire the future research on VQA or other multi-modal tasks, such as image captioning, with missing modalities~\cite{kim2019image}.

\vfill